\documentclass[10pt, a4paper, conference, compsocconf]{IEEEtran}
\ifCLASSINFOpdf
\else
\fi
\usepackage{calligra}
\usepackage{graphicx}
\usepackage{amsfonts}

\usepackage[usenames, dvipsnames]{color}

\usepackage{float}
\usepackage{mathtools}          

\floatstyle{plaintop}\restylefloat{table} 

\usepackage{booktabs} 
\usepackage{makecell} 
\usepackage{relsize}  
\usepackage{url}

\usepackage{ifthen}
\newcommand\myfootnote[2][]{\ifthenelse{\equal{#1}{}}{\footnote{#2}}{\footnote{\label{#1}#2}}}
\makeatletter
\newcommand\myfootnotemark[1]{\protected@xdef\@thefnmark{\ref{#1}}\@footnotemark}
\makeatother

\renewenvironment{thebibliography}[1]{
  \begin{oldthebibliography}{#1}
    \setlength{\itemsep}{0em}
    \setlength{\parskip}{0.45em}
}
{
  \end{oldthebibliography}
}

\begin{document}
%
\title{Joint Recognition of Handwritten Text and Named Entities with a Neural End-to-end Model}




%
\author{\IEEEauthorblockN{Manuel Carbonell\IEEEauthorrefmark{1}\IEEEauthorrefmark{2},
Mauricio Villegas\IEEEauthorrefmark{1},
Alicia Forn\'{e}s\IEEEauthorrefmark{2} and
Josep Llad\'{o}s\IEEEauthorrefmark{2}}
\IEEEauthorblockA{\IEEEauthorrefmark{1}
omni:us\\
Berlin, Germany,\\
\{manuel,mauricio\}@omnius.com}
\IEEEauthorblockA{\IEEEauthorrefmark{2}
Computer Vision Center -  Computer Science Department\\
Universitat Autonoma de Barcelona, Spain\\
\{afornes,josep\}@cvc.uab.es}
}


\maketitle

\begin{abstract}

When extracting information from handwritten documents, text transcription and named entity recognition are usually faced as separate subsequent tasks. This has the disadvantage that errors in the first module affect heavily the performance of the second module. In this work we propose to do both tasks jointly, using a single neural network with a common architecture used for plain text recognition.
Experimentally, the work has been tested on a collection of historical marriage records. Results of experiments are presented to show the effect on the performance for different configurations: different ways of encoding the information, doing or not transfer learning and processing at text line or multi-line region level.
The results are comparable to state of the art reported in the ICDAR 2017 Information Extraction competition, even though the proposed technique does not use any dictionaries, language modeling or post processing.

\end{abstract}


%
\IEEEpeerreviewmaketitle

\section{Introduction}




Extracting information from historical handwritten text documents in an optimal and efficient way is still a challenge to solve, since text in these kind of documents are not as simple to read as printed characters or modern handwritten calligraphies \cite{VeronicaRomero2016}, \cite{Toselli:2016:HWG:3043320.3051195}. Historical manuscripts contain information that gives an interpretation of the past of societies. Systems designed to search and retrieve information from historical documents must go beyond literal transcription of sources. Indeed it is necessary to shorten the semantic gap and get semantic meaning from the contents, thus the extraction of the relevant information carried out by named entities (e.g. names of persons, organizations, locations, dates, quantities, monetary values, etc.) is a key component of such systems. Semantic annotation of documents, and in particular automatic named entity recognition is neither a perfectly solved problem \cite{DBLP:journals/corr/LampleBSKD16}.

Many existing solutions make use of Artificial Neural Networks (ANNs) to transcribe handwritten text lines and then parse the transcribed text with a Named Entity Recognition model, but the precision of those existing solutions is still to improve \cite{VeronicaRomero2016}, \cite{Toselli:2016:HWG:3043320.3051195}, \cite{competition}. One possible approach is to start with already segmented words, by an automatic or manual process, and predict the semantic category using visual descriptors (c.f. \cite{Toledo2016}) which has the benefit that when the name entity prediction is correct, the transcription would be much easier to predict correctly since it restricts the language model within the corresponding category. The downside is that we rarely have large amounts of word level segmented data, a key for most ANNs proper performance. In case that automatic word segmentation is needed, the whole information extraction process involves three steps which will probably accumulate errors in each of them.
Another and most common option is to perform handwritten text recognition (HTR) first and then named entity recognition (NER). An advantage of this approach is that it has one less step than the previous explained approach, but it has the counterpart that if the transcription is wrong, the NER part is affected.

Recent work in ANNs suggests that using models that solve tasks as general as possible, might give similar or better performance than concatenating subprocesses due to error propagation in the different steps, as shown in \cite{DBLP:journals/corr/LiuFQJY16}, \cite{DBLP:journals/corr/BojarskiTDFFGJM16}. This is the main motivation of this work, and consequently we propose a single convolutional-sequential model to jointly perform transcription and semantic annotation. Adding a language model, the transcription can be restricted to each semantic category and therefore improved. The contribution of this work is to show the improvement when joining a sequence of processes in a single one, and thus, avoiding to commit accumulation of errors and achieving generalization to emulate human-like intelligence.

Some examples of historical handwritten text documents include birth, marriage and defunction records which provide very meaningful information to reconstruct genealogical trees and track locations of family ancestors, as well as give interesting macro-indicators to scholars in social sciences and humanities. The interpretation of such types of documents unavoidably requires the identification of named entities. As experimental scenario we illustrate the performance of the proposed method on a collection of handwritten marriage records.

The rest of the paper is organized as follows: Next section explains the task being considered. In section \ref{stateofart} we review the state of the art work in HTR and NER. In \ref{methodology} we explain our model architecture, ground truth setup and training details. In Section \ref{results} we analyze the results for the different configurations and last in \ref{conclusion} we give the conclusions.

\begin{figure*}
\centering
\includegraphics[width=0.95\textwidth]{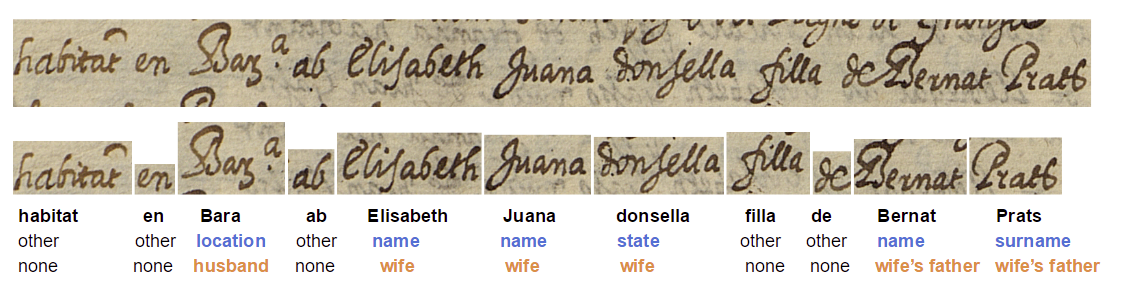}
\caption{\label{fig:frog}An example of a document line annotation from \cite{competition}.}
\label{competit_annotation}
\end{figure*}

\section{The Task: Information Extraction in Marriage Records}
\label{task}

The approach presented in this paper is general enough to be applied to many information extraction tasks, but due to time constraints and our access to a particular dataset, the approach is evaluated on the task of information extraction in a system for the analysis of population records, in particular handwritten marriage records. It consists of transcribing the text and to assign to each word a semantic and person category, i.e. to know which kind of word has been transcribed (name, surname, location, etc.) and to what person it refers to. The dataset and evaluation protocol are exactly the same as the one proposed in the ICDAR 2017 Information Extraction from Historical Handwritten Records (IEHHR) competition \cite{competition}.
%
The semantic and person categories to identify in the IEHHR competition are listed in table \ref{categories}.

\begin{table}
\centering
\caption{Semantic and person categories in the IEHHR competition}
\label{categories}
\begin{tabular}{ll}
\textbf{Semantic} & \textbf{Person}                           \\ \hline
Name              & Wife                                      \\
Surname           & Husband                                   \\
Occupation        & Wife's father                             \\
Location          & Wife's Mother                             \\
Civil State       & Husband's father                          \\
Other             & Husband's mother                          \\
                  & Other person \\
                  & None
\end{tabular}
\end{table}

Two tracks were proposed. In the basic track the goal is to assign the semantic class to each word, whereas in the complete track it is also necessary to identify the person. An example of both tracks is shown in Figure \ref{competit_annotation}.

The dataset for this competition contains 125 pages with 1221 marriage records (paragraphs), where each record contains several text lines giving information of the wife, husband and their parents' names, occupations, locations and civil states. The text images are provided at word and line level, naturally having the increased difficulty of word segmentation when choosing to work with line images. More details of the dataset can be found in table \ref{dataset_details}.

\begin{table}
\centering
\caption{Marriage Records dataset distribution}
\label{dataset_details}
\begin{tabular}{llll}
        & Train & Validation & Test \\ \hline
Pages   & 90    & 10         & 25   \\
Records & 872   & 96         & 253  \\
Lines   & 2759  & 311        & 757  \\
Words   & 28346 & 3155       & 8026 \\ \hline
\multicolumn{4}{l}{Out of vocabulary words: 5.57 \%}
\end{tabular}
\end{table}

\section{State of the art}
\label{stateofart}
Recent work shows that neural models allow generalization of problems that earlier were solved separately \cite{DBLP:journals/corr/BojarskiTDFFGJM16}. This idea can also be applied to information extraction from handwritten text documents which consists of HTR followed by NER. From the HTR side there is still a long way to improve until human level transcription is achieved \cite{NIPS2016_6257}. Attention models have helped to understand the inside behavior of neural networks when reading document images but still have lower accuracy than Recurrent Neural Network with Connectionist Temporal Classification (RNN+CTC) approaches \cite{DBLP:journals/corr/BlucheLM16}.

Named entity recognition is the problem of detecting and assigning a category to each word in a text, either at part-of-speech level or in pre-defined categories such as the names of persons, organizations, locations, expressions of times, quantities, monetary values, percentages, etc. The goal is to select and parse relevant information from the text and relationships within it. One could think that it would be sufficient to keep a list of locations, common names and organizations, but the case is that these lists are rarely complete, or one single name can refer to different kind of entities. Also it is not easy to detect properties of a named entity and how different named entities are related to each other. Most widely used kind of models for this task are \textit{conditional random fields} (CRFs), which were the state of the art technique for some time \cite{Lafferty:2001:CRF:645530.655813}, \cite{Finkel:2005:INI:1219840.1219885}.

In the area of Natural Language Processing, Lample et al. \cite{DBLP:journals/corr/LampleBSKD16} proposed a combination of \textit{Long Short-term Memory networks} (LSTMs) and CRFs, obtaining good results for the CoNLL2003 task. The problem is similar to the one we are facing, except that it starts from raw text. In this work the input to the system are images of handwritten text lines, for which it is not even known how many characters or words are present. This undoubtedly introduces a higher difficulty.

In Adak's work \cite{7490147} a similar end-to-end approach from image to semantically annotated text is proposed, but in that case the key relies in identifying capital letters to detect possible named entities. The problem is that in many cases, such as in the IEHHR competition \cite{competition} dataset, named entities do not always have capital letters, and also, it is a task-specific approach that could not be used in many other cases.

Finally, another concept that can help to improve the quality of our models' prediction is curriculum learning \cite{Bengio:2009:CL:1553374.1553380}. Letting the model look at the data in a meaningful and ordered way, such that the difficulty of prediction goes from easy to hard, and therefore, can make the training evolve with a much better performance.



\section{Methodology}
\label{methodology}

The main goal of this work is to explore a few possibilities for a single end-to-end trainable ANN model that receives as input text images and gives as output transcripts, already labeled with their corresponding semantic information. One possibility to solve it could be to propose a ANN with two sequence outputs, one for the transcript and the other for the semantic labels. However, keeping an alignment between these two independent outputs complicates a solution. An alternative would be to have a single sequence output that combines the transcript and semantic information, which is the approach taken here. There are several ways in which this information can be encoded such that a model learns to predict it. The next subsection describes the different ways of encoding it that were tested in this work. Then there are subsections describing the architecture chosen for the neural network, the image input and characteristics of the learning.

\subsection{Semantic encoding}
The first variable which we explored is the way in which ground truth transcript and semantic labels are encoded so that the model learns to predict them. To allow the model to recognize words not observed during training (out-of-vocabulary) the symbols that the model learns are the individual characters and a space to identify separation between words. For the semantic labels special tags are added to the list of symbols for the recognizer. The different possibilities are explained below.

\subsubsection{Open \& close separate tags}
In the first approach, the words are enclosed between \textbf{opening and closing tags} that encode the semantic information. Both the category and the person have independent tags. Thus, each word is encoded by starting with opening category and person symbols, followed by a symbol for each character and ends by closing person and category symbols. The ``other'' and ``none'' semantics are not encoded. For example, the ground truth of the image shown in Figure \ref{fig:frog} would be encoded as:
\begin{center}
\texttt{\item h a b i t a t \{space\} e n \{space\} <location> <husband> B a r a </husband> </location> \{space\} a b \{space\} <name> <wife> E l i s a b e t h  </wife> </name> ... }
\end{center}

This kind of encoding is not expected to perform well in the IEHHR task, since tags are assigned to only one word at a time, so it is redundant to have two tags for each word. However, in other tasks it could make sense having opening and closing tags and this is why it has been considered in this work.

\subsubsection{Single separate tags}
Similar to the previous approach, in this case both category and person tags are independent symbols but there is only one for each word added before the word. Thus, the ground truth of the previous example would be encoded as:
\begin{center}
\texttt{\item h a b i t a t \{space\} e n \{space\} <location/> <husband/> B a r a \{space\} a b \{space\} <name/> <wife/> E l i s a b e t h \{space\} J u a n a \{space\} <state/> <wife/> \{space\} d o n s e l l a ... }

\end{center}
\subsubsection{Change of person tag}
\begin{figure}
\includegraphics[width=0.5\textwidth]{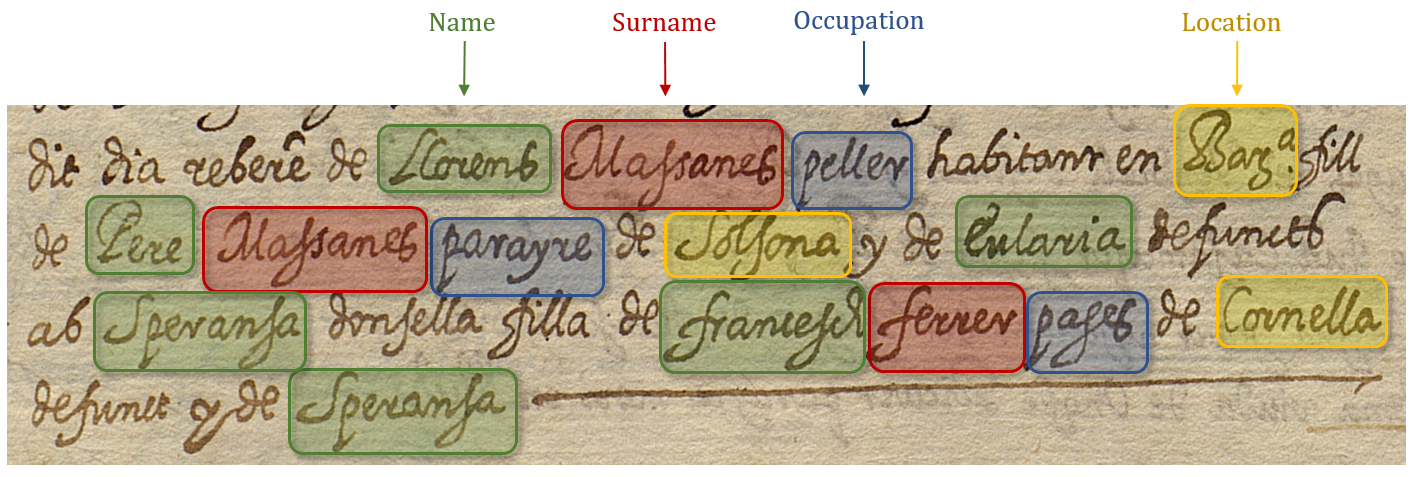}
\label{iehhr_record}
\caption{Reading the whole record makes it easier to transcribe as well as to identify the semantic categories based on context information.}
\label{fig:iehhr_record}
\end{figure}

In this variation of the semantic encoding the person label is only given if there is a \textbf{change of person}, i.e. the person label indicates that all the upcoming words refer to that person until another person label comes, in contrast to previous approaches where we give the person label for each word. This approach is possible due to the structured form of the sentences in the dataset. As we can see in Figure \ref{fig:iehhr_record} the marriage records give the information of all the family members without mixing them.
\begin{center}
\texttt{<wife/> <name/> E l i s a b e t h \{space\} <name/> J u a n a \{space\} <state/> d o n s e l l a ... }
\end{center}

\subsubsection{Single combined tags}
The final possibility tested for encoding the named entity information is to \textbf{combine category and person} labels into a single tag. So the example would be as:
\begin{center}
\texttt{ h a b i t a t \{space\} e n \{space\} <location\_husband/> B a r a \{space\} a b \{space\} <name\_wife/> E l i s a b e t h \{space\} <name\_wife/> J u a n a \{space\} <state\_wife/> d o n s e l l a ... }
\end{center}

\subsection{Level of input images: lines or records}
The IEHHR competition dataset includes manually segmented images at word level. But to lower ground truthing cost or avoid needing a word segmentator, we will assume that only images at line level are available. Having text line images then the obvious approach is to give the system individual line images for recognition. However, there are semantic labels that would be very difficult to predict if only a single line image is observed due to lack of context. For example, it might be hard to know if the name of a person corresponds to the husband or the father of the wife if the full record is not given. Because of this, in the experiments we have explored having as input both text line images and full marriage record images, concatenating all the lines of a record one after the other.



\subsection{Transfer learning}
The next variable we examined was the effect of the use of \textbf{transfer learning} from a previously trained model for HTR. Transfer Learning consists of training for the same or a similar task (HTR) using other datasets, and then fine tune it for our purpose, in our case HTR+NER. To perform transfer learning from a generic HTR model, the softmax layer is removed and replaced with a softmax that allows as an output the activations for the number of possible classes in the fine tuning step. In our case, they will be all the characters in the alphabet plus the semantic labels. In the experiments for transfer learning we have tested only one HTR model that was trained with the following datasets: IAM~\cite{Marti99afull}, Bentham~\cite{6981116}, Bozen~\cite{Sanchez2016a}, and some datasets used by us internally: IntoThePast, Wiensanktulrich, Wienvotivkirche and ITS.

\subsection{Curriculum Learning}
The last variation that we propose is curriculum learning i.e. start with easier demands to the model and then increase the difficulty. In this case this method can be interpreted as starting by learning to transcribe single text lines, and when the training is finished, continue with learning to transcribe images of a whole marriage record.

\subsection{Model architecture and training}
\begin{figure*}
\centering
\includegraphics[scale=0.45]{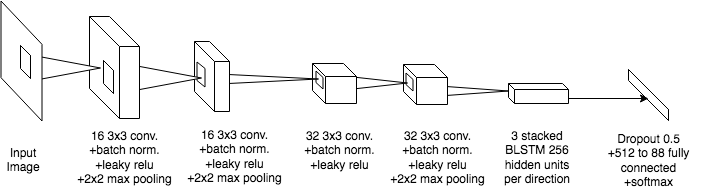}
\caption{Used model architecture}
\label{architecture}
\end{figure*}

In this work we use a CNN+BLSTM+CTC model, which is one of the most common models for performing HTR exclusively, although other HTR models could be used as well. In particular, the architecture consists of 4 convolutional layers with max pooling followed by 3 stacked BLSTM layers. The detailed model architecture is shown in Figure \ref{architecture}.

To train the model we use the Laia HTR toolkit \cite{laia2016} which uses Baidu's parallel CTC \cite{Graves:2006:CTC:1143844.1143891} implementation, which consists of minimizing the loss or ``objective'' function
\begin{equation}
O^{ML}(S,\mathcal{N}_w)=-\sum_{(x,z)\in S} ln(p(z|x))
\end{equation}
where $S$ is the training set, $x$ is the input sequence (visual features), $z$ is the sequence labeling (transcription) for $x$ and
\begin{equation}
\mathcal{N}_w:(\mathbb{R}^m)^T\mapsto(\mathbb{R}^n)^T
\end{equation}
is a recurrent neural network with $m$ inputs, $n$ outputs and weight vector $w$. The probabilities of a labeling of an input sequence are calculated with a dynamic programming algorithm called "forward-backward".



Some special features of our model are that the activation function for the convolutional layers is leaky ReLu
 $f(x) =  x \ \text{if $x>0.01$},\ 0.01x \text{ otherwise.}$

We also use batch normalization to reduce \textit{internal covariate shift}~\cite{DBLP:journals/corr/IoffeS15}.

\section{Results}
\label{results}
We compare the performance of our methods%
\footnote{Scripts used for the experiments available at \url{http://doi.org/10.5281/zenodo.1174113}}
with the results of the participants of the IEHHR competition in \cite{competition} thereby using the same metric, see Table \ref{results_table}. The evaluation metric counts the words that were correctly transcribed and annotated with their category and person label with respect to the total amount of words in the ground truth. For those words that were not correctly transcribed but the category and person labels match one or more words in the ground truth, we add to the score 1 - CER (character error rate) on the best matching word. This means that the named entity recognition part is vital for a good score, since a perfect transcription will count as 0 in the score if its named entity is incorrectly detected.

\begin{table}
  \caption{Average scores of the experiments compared with the IEHHR competition participants' methods.\label{tab:results}}
  \label{results_table}
  \centering
  \begin{minipage}{\linewidth}
  \relsize{-1}
  \newcommand{\HS}{\hspace{5pt}}
  \newcommand{\LS}{\addlinespace[4pt]}
  \begin{tabular}{ccccc}
    \toprule
    {\bf Method} & \makecell{\bf Segm. \\\bf Level} & \makecell{\bf Proc. \\\bf Level} & \makecell{\bf Track \\\bf Basic} & \makecell{\bf Track \\\bf Complete} \\
    \midrule
    \multicolumn{5}{c}{\relsize{-1}\bf IEHHR competition results} \\
    \midrule
    \makecell[l]{Hitsz-ICRC-1 \\\HS CNN HTR+NER}       & Word       & Record\myfootnote[fn:hitsz]{\relsize{-1}HTR is word based.}   & 87.56 & 85.72 \\\LS
     \makecell[l]{Hitsz-ICRC-2 \\\HS ResNet HTR+NER}    & Word          & Record\myfootnotemark{fn:hitsz}   & 94.16 & 91.97 \\\LS

    \makecell[l]{Baseline \\\HS HMM+MGGI}              & Line & Record   & 80.24 & 63.08 \\\LS
    \makecell[l]{CITlab-ARGUS-1 \\\HS LSTM+CTC+regex}     & Line  & Record\myfootnote[fn:citlab]{\relsize{-1}Posterior character probabilities computed at line level.}
    & 89.53 & 89.16 \\\LS
    \makecell[l]{CITlab-ARGUS-2 \\\HS LSTM+CTC \\\HS +OOV+regex} & Line  & Record\myfootnotemark{fn:citlab}
    &\textbf{ 91.93 }& \textbf{91.56} \\
    \midrule
    \multicolumn{5}{c}{\relsize{-1}\bf Results of our experiments} \\
    \midrule
    \makecell[l]{Separate-single tags }
      & Line & Line
      & 73.49
      & 61.96 \\\LS
    \makecell[l]{Separate-\\\HS open-close tags}
      & Line & Line
      & 73.70
      & 64.09 \\\LS

   \makecell[l]{Combined-single tags}
      & Line & Line
      & 87.96
      & 80.74 \\\LS

    \makecell[l]{Combined-single tags \\\HS + transfer learn}
      & Line & Line
      & 87.01
      & 80.05 \\\LS

    \makecell[l]{Change person tag \\\HS + transfer learn}
      & Line & Record
      & 84.41
      & 80.51 \\\LS

    \makecell[l]{Combined-single tags \\\HS + transfer learn}
      & Line & Record
      & 86.58
      & 84.72 \\\LS

    \makecell[l]{Combined-single tags \\\HS + transfer learn \\\HS + curriculum learn}
      & Line & Record
      & \textbf{90.58}
      & \textbf{89.39} \\\LS
    \makecell[l]{Combined-single tags \\\HS + transfer learn \\\HS + curriculum learn \\\HS + alt. line extraction}
    & Word\myfootnote[fn:comp]{\relsize{-1}Not fair to compare with the IEHHR results because it uses a different segmentation (alternative line extraction) than the one provided in the competition.}
    &\makecell[l]{Record\myfootnotemark{fn:comp}}
      & \textbf{96.39}\myfootnotemark{fn:comp}
      & \textbf{96.63}\myfootnotemark{fn:comp} \\\LS
    \bottomrule
  \end{tabular}
  \end{minipage}
\end{table}

We can observe in the results that our best performance is reached when receiving the whole marriage record, which is probably due to the help of contextual information. For example, it can benefit the detection of named entities composed of several words when they are written in separate consecutive lines. Also we observe that the best performing encoding of the semantic labels is the combined tags setup. This can be due to the lower amount of symbols to predict, which might require to store less long term dependencies in the network.

The most significant improvement was achieved when picking our best performing configuration and running it with an alternative line extraction. In the competition, the text lines were extracted by including all the bounding boxes of the words within every line. As a result, when there are large ascenders and descenders, the bounding box of the line is too wide, including sections of other text lines. In order to cope with this limitation, we used the XML containing the exact location of the segmented words within a page, and for the y-coordinates, we used a weighted (by the words widths) average of upper and lower limits of the word bounding boxes. As expected, the performance highly improves because the segmentation of the text lines is more accurate. However, this result is not directly comparable to the other participants's methods because the segmentation is different.


\begin{figure*}
\centering
\includegraphics[width=0.8\textwidth]{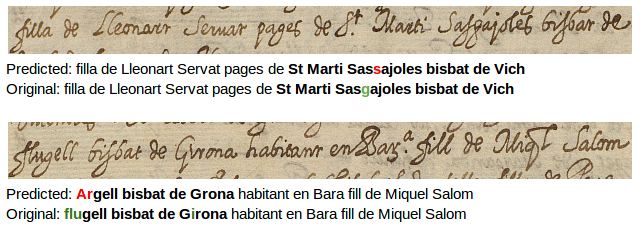}
\caption{Some of the errors committed in the predictions}
\label{fig_error1}
\end{figure*}

In Figure \ref{fig_error1} we show some examples of committed errors. We can see that they consist of small typos that are understandable when looking at the text images. It is definitely difficult to transcribe certain names that have never been seen before. The proposed approach could be combined with a category-based language model~\cite{VeronicaRomero2016} which could potentially improve the results.

\begin{figure}
\includegraphics[width=0.5\textwidth]{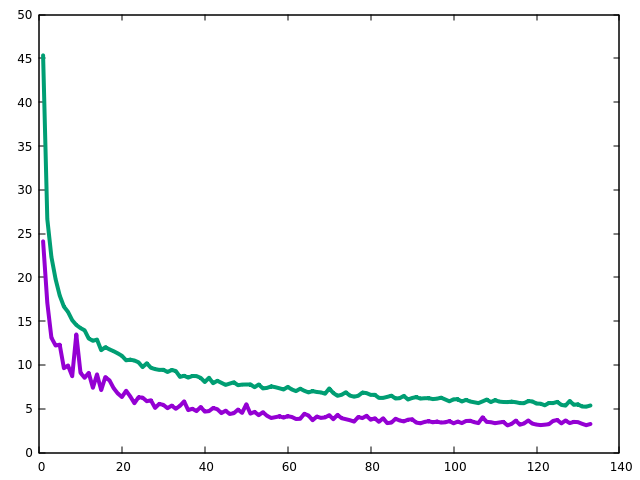}
\caption{Train and validation (green and violet respectively) CER (\%).}
\label{fig_train}
\end{figure}

Our best performing model took 4 hours 38 to run 133 training epochs with a NVIDIA GTX 1080 GPU. The train and validation error rates can be seen in Figure \ref{fig_train}. As training configuration we used an adversarial regularizer~\cite{Goodfellow14_arXiv} with weight 0.5, an initial learning rate of $5\cdot 10^{-4}$ with decay factor of 0.99 per epoch and batch size 6.

\section{Conclusion}
\label{conclusion}
In this paper we have proposed to solve a complex task (i.e. text recognition and named entity recognition) with a single end-to-end neural model. Our first conclusion is that, also in information extraction problems, a generic model for solving two subsequent tasks can perform at least similarly as two separated models. This is true even if there is less prepared data (record level images instead of a sequence of word images) and we do not make use of task specific tools like dictionaries or language model. 

By investigating different ways of encoding the image transcripts and semantic labels we have shown that the recognition performance is highly affected, even though it is indeed representing the same information. Also, curriculum learning (first text lines and then records) can make the model reach a higher final prediction accuracy.

Future work would include the use of language models to improve the accuracy of the predictions, the effect of automatic text line and record detection, and also, to evaluate our method in other datasets.

\section*{Acknowledgments}

This work has been partially supported by the Spanish project TIN2015-70924-C2-2-R, the grant 2016-DI-095 from the Secretaria d'Universitats i Recerca del Departament d'Economia i Coneixement de la Generalitat de Catalunya, the Ramon y Cajal Fellowship RYC-2014-16831, the CERCA Programme /Generalitat de Catalunya, and RecerCaixa (XARXES, 2016ACUP-00008), a research program from Obra Social "La Caixa" with the collaboration of the ACUP.



%

\bibliographystyle{IEEEtran}
\bibliography{IEEEabrv,sample}

\begin{thebibliography}{10}
\providecommand{\url}[1]{#1}
\csname url@samestyle\endcsname
\providecommand{\newblock}{\relax}
\providecommand{\bibinfo}[2]{#2}
\providecommand{\BIBentrySTDinterwordspacing}{\spaceskip=0pt\relax}
\providecommand{\BIBentryALTinterwordstretchfactor}{4}
\providecommand{\BIBentryALTinterwordspacing}{\spaceskip=\fontdimen2\font plus
\BIBentryALTinterwordstretchfactor\fontdimen3\font minus
  \fontdimen4\font\relax}
\providecommand{\BIBforeignlanguage}[2]{{%
\expandafter\ifx\csname l@#1\endcsname\relax
\typeout{** WARNING: IEEEtran.bst: No hyphenation pattern has been}%
\typeout{** loaded for the language `#1'. Using the pattern for}%
\typeout{** the default language instead.}%
\else
\language=\csname l@#1\endcsname
\fi
#2}}
\providecommand{\BIBdecl}{\relax}
\BIBdecl

\bibitem{VeronicaRomero2016}
V.~Romero, A.~Fornes, E.~Vidal, and J.~A. Sanchez, ``Using the mggi methodology
  for category-based language modeling in handwritten marriage licenses
  books,'' in \emph{15th international conference on Frontiers in Handwriting
  Recognition}, 2016.

\bibitem{Toselli:2016:HWG:3043320.3051195}
\BIBentryALTinterwordspacing
A.~H. Toselli, E.~Vidal, V.~Romero, and V.~Frinken, ``Hmm word graph based
  keyword spotting in handwritten document images,'' \emph{Inf. Sci.}, vol.
  370, no.~C, pp. 497--518, Nov. 2016. [Online]. Available:
  \url{https://doi.org/10.1016/j.ins.2016.07.063}
\BIBentrySTDinterwordspacing

\bibitem{DBLP:journals/corr/LampleBSKD16}
\BIBentryALTinterwordspacing
G.~Lample, M.~Ballesteros, S.~Subramanian, K.~Kawakami, and C.~Dyer, ``Neural
  architectures for named entity recognition,'' \emph{CoRR}, vol.
  abs/1603.01360, 2016. [Online]. Available:
  \url{http://arxiv.org/abs/1603.01360}
\BIBentrySTDinterwordspacing

\bibitem{competition}
A.~Forn{\' e}s, V.~Romero, A.~Bar{\'o}, J.~I. Toledo, J.~A. Sanchez, E.~Vidal,
  and J.~Llad{\'o}s, ``Competition on information extraction in historical
  handwritten records,'' in \emph{International Conference on Document Analysis
  and Recognition (ICDAR)}.\hskip 1em plus 0.5em minus 0.4em\relax IEEE, 2017.

\bibitem{Toledo2016}
J.~I. Toledo, S.~Sudholt, A.~Forn{\'e}s, J.~Cucurull, G.~A. Fink, and
  J.~Llad{\'o}s, \emph{Handwritten Word Image Categorization with Convolutional
  Neural Networks and Spatial Pyramid Pooling}.\hskip 1em plus 0.5em minus
  0.4em\relax Cham: Springer International Publishing, 2016, pp. 543--552.

\bibitem{DBLP:journals/corr/LiuFQJY16}
\BIBentryALTinterwordspacing
H.~Liu, J.~Feng, M.~Qi, J.~Jiang, and S.~Yan, ``End-to-end comparative
  attention networks for person re-identification,'' \emph{CoRR}, vol.
  abs/1606.04404, 2016. [Online]. Available:
  \url{http://arxiv.org/abs/1606.04404}
\BIBentrySTDinterwordspacing

\bibitem{DBLP:journals/corr/BojarskiTDFFGJM16}
\BIBentryALTinterwordspacing
M.~Bojarski, D.~D. Testa, D.~Dworakowski, B.~Firner, B.~Flepp, P.~Goyal, L.~D.
  Jackel, M.~Monfort, U.~Muller, J.~Zhang, X.~Zhang, J.~Zhao, and K.~Zieba,
  ``End to end learning for self-driving cars,'' \emph{CoRR}, vol.
  abs/1604.07316, 2016. [Online]. Available:
  \url{http://arxiv.org/abs/1604.07316}
\BIBentrySTDinterwordspacing

\bibitem{NIPS2016_6257}
T.~Bluche, ``{Joint Line Segmentation and Transcription for End-to-End
  Handwritten Paragraph Recognition},'' in \emph{Advances in Neural Information
  Processing Systems 29}, D.~D. Lee, M.~Sugiyama, U.~V. Luxburg, I.~Guyon, and
  R.~Garnett, Eds.\hskip 1em plus 0.5em minus 0.4em\relax Curran Associates,
  Inc., 2016, pp. 838--846.

\bibitem{DBLP:journals/corr/BlucheLM16}
\BIBentryALTinterwordspacing
T.~Bluche, J.~Louradour, and R.~O. Messina, ``Scan, attend and read: End-to-end
  handwritten paragraph recognition with {MDLSTM} attention,'' \emph{CoRR},
  vol. abs/1604.03286, 2016. [Online]. Available:
  \url{http://arxiv.org/abs/1604.03286}
\BIBentrySTDinterwordspacing

\bibitem{Lafferty:2001:CRF:645530.655813}
\BIBentryALTinterwordspacing
J.~D. Lafferty, A.~McCallum, and F.~C.~N. Pereira, ``Conditional random fields:
  Probabilistic models for segmenting and labeling sequence data,'' in
  \emph{Proceedings of the Eighteenth International Conference on Machine
  Learning}, ser. ICML '01.\hskip 1em plus 0.5em minus 0.4em\relax San
  Francisco, CA, USA: Morgan Kaufmann Publishers Inc., 2001, pp. 282--289.
  [Online]. Available: \url{http://dl.acm.org/citation.cfm?id=645530.655813}
\BIBentrySTDinterwordspacing

\bibitem{Finkel:2005:INI:1219840.1219885}
\BIBentryALTinterwordspacing
J.~R. Finkel, T.~Grenager, and C.~Manning, ``Incorporating non-local
  information into information extraction systems by gibbs sampling,'' in
  \emph{Proceedings of the 43rd Annual Meeting on Association for Computational
  Linguistics}, ser. ACL '05.\hskip 1em plus 0.5em minus 0.4em\relax
  Stroudsburg, PA, USA: Association for Computational Linguistics, 2005, pp.
  363--370. [Online]. Available: \url{https://doi.org/10.3115/1219840.1219885}
\BIBentrySTDinterwordspacing

\bibitem{7490147}
C.~Adak, B.~B. Chaudhuri, and M.~Blumenstein, ``Named entity recognition from
  unstructured handwritten document images,'' in \emph{2016 12th IAPR Workshop
  on Document Analysis Systems (DAS)}, April 2016, pp. 375--380.

\bibitem{Bengio:2009:CL:1553374.1553380}
\BIBentryALTinterwordspacing
Y.~Bengio, J.~Louradour, R.~Collobert, and J.~Weston, ``Curriculum learning,''
  in \emph{Proceedings of the 26th Annual International Conference on Machine
  Learning}, ser. ICML '09.\hskip 1em plus 0.5em minus 0.4em\relax New York,
  NY, USA: ACM, 2009, pp. 41--48. [Online]. Available:
  \url{http://doi.acm.org/10.1145/1553374.1553380}
\BIBentrySTDinterwordspacing

\bibitem{Marti99afull}
U.~v.~Marti and H.~Bunke, ``A full english sentence database for off-line
  handwriting recognition,'' in \emph{In Proc. Int. Conf. on Document Analysis
  and Recognition}, 1999, pp. 705--708.

\bibitem{6981116}
J.~A. S{\'a}nchez, V.~Romero, A.~H. Toselli, and E.~Vidal, ``Icfhr2014
  competition on handwritten text recognition on transcriptorium datasets
  (htrts),'' in \emph{2014 14th International Conference on Frontiers in
  Handwriting Recognition}, Sept 2014, pp. 785--790.

\bibitem{Sanchez2016a}
J.~S{\'a}nchez, V.~Romero, A.~Toselli, and E.~Vidal, ``{ICFHR}2016 competition
  on handwritten text recognition on the {READ} dataset,'' in
  \emph{ICFHR}.\hskip 1em plus 0.5em minus 0.4em\relax IEEE, 2016, pp.
  630--635.

\bibitem{laia2016}
J.~Puigcerver, D.~Martin-Albo, and M.~Villegas, ``Laia: A deep learning toolkit
  for htr.''\hskip 1em plus 0.5em minus 0.4em\relax GitHub, 2016, gitHub
  repository.

\bibitem{Graves:2006:CTC:1143844.1143891}
\BIBentryALTinterwordspacing
A.~Graves, S.~Fern\'{a}ndez, F.~Gomez, and J.~Schmidhuber, ``Connectionist
  temporal classification: Labelling unsegmented sequence data with recurrent
  neural networks,'' in \emph{Proceedings of the 23rd International Conference
  on Machine Learning}, ser. ICML '06.\hskip 1em plus 0.5em minus 0.4em\relax
  New York, NY, USA: ACM, 2006, pp. 369--376. [Online]. Available:
  \url{http://doi.acm.org/10.1145/1143844.1143891}
\BIBentrySTDinterwordspacing

\bibitem{DBLP:journals/corr/IoffeS15}
\BIBentryALTinterwordspacing
S.~Ioffe and C.~Szegedy, ``Batch normalization: Accelerating deep network
  training by reducing internal covariate shift,'' \emph{CoRR}, vol.
  abs/1502.03167, 2015. [Online]. Available:
  \url{http://arxiv.org/abs/1502.03167}
\BIBentrySTDinterwordspacing

\bibitem{Goodfellow14_arXiv}
I.~J. Goodfellow, J.~Shlens, and C.~Szegedy, ``Explaining and harnessing
  adversarial examples,'' 2014.

\end{thebibliography}

\end{document}